\documentclass{article}

\usepackage[final]{corl_2019} % Uncomment for the camera-ready ``final'' version
\usepackage{amsmath,amssymb,amsbsy}
\usepackage{xfrac}
\usepackage{graphicx}
\usepackage{subcaption}
\usepackage{booktabs}
\usepackage{makecell}
\usepackage{multicol}
\usepackage{multirow}
\usepackage{verbatim}
\usepackage{xcolor}
\usepackage{wrapfig}

\DeclareMathOperator*{\argtopk}{arg\_top\_k}

\title{Leveraging Semantics for Incremental Learning \\ in Multi-Relational Embeddings}

\author{
  Angel Daruna, Weiyu Liu, Zsolt Kira, Sonia Chernova \\
  Institute for Robotics and Intelligent Machines\\
  Georgia Institute of Technology, United States\\
  \texttt{\{adaruna3,wliu88,zkira,chernova\}@gatech.edu} \\
}

\begin{document}
\maketitle

%===============================================================================

\begin{abstract}
Service robots benefit from encoding information in semantically meaningful ways to enable more robust task execution. Prior work has shown multi-relational embeddings can encode semantic knowledge graphs to promote generalizability and scalability, but only within a batched learning paradigm. We present \textit{Incremental Semantic Initialization (ISI)}, an incremental learning approach that enables novel semantic concepts to be initialized in the embedding in relation to previously learned embeddings of semantically similar concepts. We evaluate ISI on mined AI2Thor and MatterPort3D datasets; our experiments show that on average ISI improves immediate query performance by 41.4\%. Additionally, ISI methods on average reduced the number of epochs required to approach model convergence by 78.2\%.
\end{abstract}

\keywords{relational learning, incremental learning, semantic reasoning} 

%===============================================================================
\vspace{-0.3cm}
\section{Introduction}
\label{sec:intro}
\vspace{-0.3cm}

Robots operating in human environments benefit from using knowledge representations that encode information in semantically meaningful ways to facilitate generalization and adaptability, leading to more robust task execution~\cite{pronobis2012large,zhu2014reasoning,beetz2018know}. An explicit commonly used model of environment semantics defines a set of entities $\mathcal{E}$ representing known concepts (e.g. apple, metal, open), and a set of possible relations $\mathcal{R}$ (e.g. atLocation, hasAffordance) between them~\cite{beetz2018know,chernovasituated,saxena2014robobrain,zhu2014reasoning}. Combined, $\mathcal{E}$ and $\mathcal{R}$ form a knowledge graph $\mathcal{G}$, in which vertices represent entities and edges represent relations.

Multiple techniques have been proposed for effectively performing inference over semantic knowledge bases.  Most recently, Daruna et al.~\cite{daruna2019robocse} showed that representing knowledge graphs using multi-relational embeddings significantly outperforms prior approaches, such as directed graphs \cite{saxena2014robobrain}, Bayesian Logic Networks \cite{chernovasituated}, and Description Logics \cite{tenorth2010knowrob}, with respect to scalability, robustness to uncertainty, and generalizability.  Multi-relational embeddings represent knowledge graphs in vector space, encoding vertices that represent entities $\mathcal{E}$ as vectors and edges that represent relations $\mathcal{R}$ as mappings. However, Daruna et al.'s work assumes all entities and relations to be known before learning the representation. This assumption is impractical for large-scale and long-term deployments of autonomous systems because each incremental discovery of a new concept would require batch retraining of the encoding.

In this work, we introduce a novel incremental learning approach for semantic data within multi-relational embeddings. We consider the ``Incremental Class Learning'' scenario~\cite{hsu2018re}, which applies to systems in which knowledge is acquired incrementally over time. Our experiments model the service robot scenario in which a home robot incrementally gains knowledge of new concepts, such as discovering new affordances, detecting new materials, or finding new objects. Our objective is to integrate each new concept into the robot's existing semantic knowledge representation as quickly and accurately as possible, while mitigating corruption to previous concepts.

The core contribution of our work -- \textit{Incremental Semantic Initialization (ISI)}\footnote{Code available at \textit{[URL withheld for blind review]}} -- enables embeddings for a novel concept (e.g. $apple$) to be initialized in relation to previously learned embeddings of semantically similar concepts (e.g. $banana$) and away from dissimilar concepts (e.g. $lamp$).  We present three variants of our approach: Entity Similarity (ES), Relational Similarity (RS), and Hybrid Similarity (ERS) that inform the initialization of new concepts using entities, relations between entities, and both, respectively. 

We validated our approach on knowledge graphs mined from AI2Thor and MatterPort3D\footnote{Datasets available at \textit{[URL withheld for blind review]}}.  Our results show that ISI significantly outperforms the state-of-the-art in incremental multi-relational embedding initialization~ \cite{song2018enriching} due to ISI's ability to initialize novel concepts in a semantically meaningful way without retraining.  Additionally, we show that all ISI methods reduce the number of epochs required to reach within 8 MRR* percentage points of a model trained with batch learning over all available data by 78.2\% on average when compared to prior work. As a result, our approach provides a significant efficiency improvement for the deployment of multi-relational embeddings onto robot systems in incremental learning scenarios.

%Previous work has reposed the multi-relational embedding problem to enable incremental learning~\cite{song2018enriching}. However, the method of initializing embeddings for new concepts used in the current state-of-art (e.g.~\cite{song2018enriching}) during incremental learning phases is problematic as it results in semantically meaningless initializations that corrupt inferences related to previous concepts and does not take advantage of previously learned concepts, leading to longer convergence times.

%===============================================================================
\vspace{-0.3cm}
\section{Related Work}
\label{sec:related}
\vspace{-0.3cm}

\textbf{Semantic Reasoning} for robotics applications commonly uses an explicit model of world semantics in which a knowledge graph $\mathcal{G}$ is composed of individual positive example facts, or triples, $(h,r,t)$ such that $h,t \in \mathcal{E}$ are identified as head and tail entities of the triple, respectively, for which the relation $r \in \mathcal{R}$ holds (e.g. {\small$($\textit{cup, hasAffordance, fill}$)$})~\cite{beetz2018know,chernovasituated,saxena2014robobrain,zhu2014reasoning}.  Multiple computational frameworks have been proposed that enable robots to reason about semantic knowledge \cite{beetz2018know,chernovasituated,zhu2014reasoning}.  Our work focuses on the recent work on multi-relational embeddings presented in \cite{daruna2019robocse}, which was shown to ourperform prior methods with respect to scalability and generalizability on batch learning tasks. 

\textbf{Multi-Relational Embeddings} model a knowledge graph $\mathcal{G}$ in vector space, encoding entities $\mathcal{E}$ as vectors and relations $\mathcal{R}$ as mappings~\cite{wang2017kge_survey}. Generically, the embeddings for $\mathcal{E}$ and $\mathcal{R}$ in $\mathcal{G}$~\footnote{Note that $\mathcal{G}$ is considered incomplete because some set of triples may be missing. Algorithms for triple classification~\cite{socher2013reasoning} and prediction (i.e. query answering) \cite{bordes2013transe} seek to account for missing information.} are learned using a scoring function $f(h,r,t)$ that maps input triples to scores so that positive triples have high scores and negative triples have low scores~\cite{nickel2016review}. As in~\cite{daruna2019robocse}, our work uses ANALOGY \cite{liu2017analogical} to learn multi-relational embeddings. ANALOGY constrains relations to be normal linear mappings between entities by using a scoring function $f(h,r,t) = \langle \textbf{v}^T_h \textbf{W}_r$,$ \textbf{v}_t \rangle$, where $\textbf{v}_h,\textbf{v}_t$ are head and tail entity vectors, respectively, and $\textbf{W}_r$ is a relation mapping. This constraint enables using far fewer parameters than the most  flexible semantic matching models~\cite{socher2013reasoning,dong2014knowledge} while allowing for more complex relations to be expressed than translational models~\cite{bordes2013transe,wang2014transh}, balancing scalability and expressiveness to achieve state-of-art results~\cite{wang2017kge_survey}. However, multi-relational embeddings assume all entities and relations to be known before training, which is impractical for robots in incremental learning scenarios.

\textbf{Continual Learning} entails learning to perform well over a new dataset or task while not degrading performance over previous datasets or tasks~\cite{hsu2018re,maltoni2018continuous,parisi2019continual}. In~\cite{hsu2018re}, continual learning is categorized by whether the distribution of input data changes-, the distribution of target labels changes-, or the labels are from a disjoint space -across learning sessions. These are referred to as `\textit{Incremental Domain Learning}', `\textit{Incremental Class Learning}', and `\textit{Incremental Task Learning}', respectively. The categorization of approaches for continual learning outlined in~\cite{maltoni2018continuous,parisi2019continual} include regularizing learning across datasets~\cite{li2018learning,kirkpatrick2017overcoming}, recalling previous dataset distributions using generative models or replay~\cite{shin2017continual,hayes2018memory}, adapting the model architecture to accommodate new datasets~\cite{rusu2016progressive,cortes2017adanet}, and using complimentary learning systems to train on new datasets~\cite{kamra2017deep}.

Previous work most related to ours~\cite{song2018enriching} reformulated the multi-relational learning objective to enable incremental learning, using \textit{normalized-initialization}~\cite{glorot2010understanding} to initialize embeddings of new concepts during learning phases. However, the normalized-initialization algorithm was developed to initialize all model weights before any training as an improvement over previous heuristics for randomly initializing all weights of neural networks. Instead of normalized-initialization, we posit that the learned embedding space should be used to inform initialization of embeddings for new concepts during incremental learning phases. Works related to this idea are that of answering out-of-knowledge-base queries.

\textbf{Out-of-Knowledge-Base (OOKB) Queries} are queries relating to concepts that are missing in a knowledge graph $\mathcal{G}$. In prior work, solutions to OOKB queries are obtained by reasoning about the current multi-relational embedding to initialize representations for OOKB concepts. In~\cite{wang2014knowledge}, the authors `align' an external knowledge source with an embedding to answer queries about OOKB concepts. In other work,~\cite{hamaguchi2017knowledge} train a graph-neural-network (GNN) to predict embeddings of OOKB concepts. The work by~\cite{shi2018open} train a deep convolutional neural network architecture to predict OOKB embeddings from text descriptions or names. We found our approach to be effective for the limited dataset size in our experiments because it requires no training to make initializations.
%===============================================================================

\vspace{-0.3cm}
\section{Problem Definition}
\label{sec:problem_def}
\vspace{-0.3cm}

The objective of the multi-relational embedding problem is to learn a continuous vector representation of a knowledge graph $\mathcal{G}$ from a dataset of triples $\mathcal{D}\!=\!\big\{(h,r,t)_i,y_i|\,h_i,t_i\!\in\!\mathcal{E},r_i\!\in\!\mathcal{R},y_i\!\in\!\{0,1\}\big\}$, in which $i\!\in\!\{1...|\mathcal{D}|\}$ and $y_i$ designates whether a relation $r_i$ holds between entities $h_i,t_i$. Each entity $e\!\in\!\mathcal{E}$ is encoded as a vector $\textbf{v}_e\!\in\!\mathbb{R}^{d_\mathcal{E}}$, and each relation $r\!\in\!\mathcal{R}$ as a mapping between vectors $\textbf{W}_r\!\in\!\mathbb{R}^{d_\mathcal{R}}$, where $d_\mathcal{E}$ and $d_\mathcal{R}$ are the dimensions of vectors and mappings, respectively~\cite{wang2017kge_survey,nickel2016review}. Therefore, the learning objective is to find a set of embeddings $\Theta = \big\{\{\textbf{v}_{e}|\,e\in\mathcal{E}\},\{\textbf{W}_{r}|\,r\in\mathcal{R}\}\big\}$ that minimize the loss over all triples in the dataset $\mathcal{L}_\mathcal{D}$; for our implementation using ANALOGY, $\mathcal{L}_\mathcal{D} = \sum_{i}-\log \sigma (y_i \cdot \langle \textbf{v}^T_{h_i} \textbf{W}_{r_i}, \textbf{v}_{t_i} \rangle)$ where $\sigma$ is a sigmoid.

The multi-relational embedding problem can be adapted for continual learning by including a new time step index $n$ that increases with each new learning session \cite{song2018enriching}. At each new learning session, the size of the entity and relation sets grow because one or more OOKB entities $\xi^{n-1}$, where $\xi^{n-1}\cap\,\mathcal{E}^{n-1}\!=\!\emptyset$, and relations $\Gamma^{n-1}$, where $\Gamma^{n-1}\cap\,\mathcal{R}^{n-1}\!=\!\emptyset$, are introduced (i.e. $\mathcal{E}^n\!=\!\mathcal{E}^{n-1}\!\cup\xi^{n-1}$ and $\mathcal{R}^n\!=\!\mathcal{R}^{n-1}\!\cup\!\Gamma^{n-1}$). Therefore, after initializing all embeddings for OOKB entities at the time step $n$, vectors for previous entities remain $\textbf{v}^n_{e}=\textbf{v}^{n-1}_{e}|\,e\in\mathcal{E}^{n-1}$ and vectors for OOKB entities are $\textbf{v}^n_{e}=\textbf{v}^n_{\hat{e}}|\,e,\hat{e}\in\xi^{n-1}$ are added, where $\textbf{v}^n_{\hat{e}}$ is generated by an OOKB entity initialization method. Embeddings for the current time step $n$ are then $\Theta^n = \big\{\{\textbf{v}^n_{e}|\,e\in\mathcal{E}^n\},\{\textbf{W}^n_{r}|\,r\in\mathcal{R}^n\}\big\}$.

As a result of incremental learning, the multi-relational embedding learning objective becomes finding a set of embeddings $\Theta^n = \big\{\{\textbf{v}^n_{e}|\,e\in\mathcal{E}^n\},\{\textbf{W}^n_{r}|\,r\in\mathcal{R}^n\}\big\}$ that minimize the loss over the dataset for that time step $\mathcal{L}_{\mathcal{D}^n}$ given the previous embeddings $\Theta^{n-1}$ and OOKB entities $\xi^{n-1}$.

%===============================================================================
\vspace{-0.3cm}
\section{Approach}
\label{sec:approach}
\vspace{-0.2cm}

\begin{wrapfigure}{r}{0.55\textwidth}
	\vspace{-1.2cm}
	\centering 
	\includegraphics[width=0.55\textwidth]{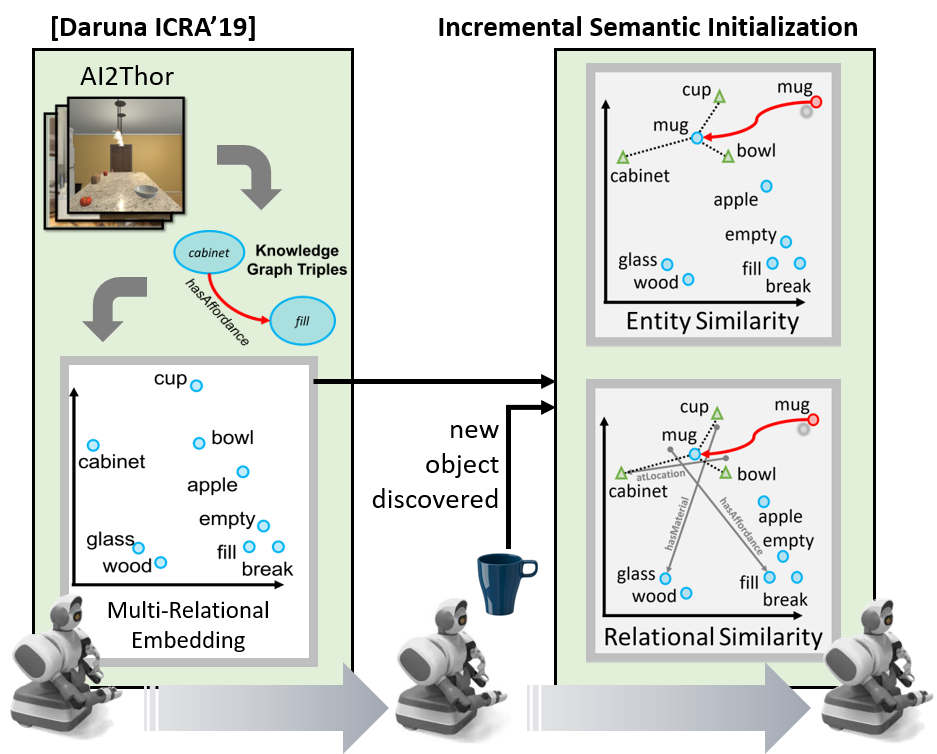}
	\caption[]
	{\small Approach overview.}
	\label{fig:approach}
	\vspace{-0.5cm}
\end{wrapfigure}
% FIGURE APPROACH
After learning a multi-relational embedding from a dataset, different regions of the entity embedding space carry distinct semantic meaning~\cite{?}.  Using normalized-initialization, as in~\cite{song2018enriching}, for new entity embeddings can severely corrupt the embedding space because initializations are void of semantic meaning as normalized-initialization was developed only to maintain activation and back-propagated gradient variances across a neural-network. We present several \textit{incremental semantic initialization (ISI)} methods for multi-relational embeddings, which reason about the learned entity embedding space to inform new entity initializations. Each method selects the most informative current entities to inform initialization of a new entity based on different embedding structure.

To initialize an embedding for an OOKB entity $\hat{e}\in\xi^{n-1}$, our algorithms rely on identifying a set of indicator entities $\mathcal{I}$ from known entities $e\in\mathcal{E}^{n-1}$. The entities in $\mathcal{I}$ indicate a reasonable region of the embedding space to initialize the OOKB entity's vector $\textbf{v}_{\hat{e}}^{n}$. In all proposed initialization algorithms, the OOKB entity vector is the centroid of the indicator entity vectors as shown below.

\vspace{-0.5cm}
\begin{equation}
\textbf{v}^{n}_{\hat{e}}=\frac{1}{|\mathcal{I}|} \sum_{e\in\mathcal{I}} \textbf{v}^{n-1}_{e}
\label{eq:indicator}
\end{equation}
\vspace{-0.5cm}

For simplicity, the algorithm descriptions below are for the case of inserting a single OOKB entity, but multiple entities can be initialized in the same time step through the same procedure.

Below we describe the three ISI methods.  Each method leverages different semantics within a multi-relational embedding to initialize OOKB entities. Entity Similarity (ES) selects indicator entities by directly comparing a new entity to current entities using word embedding similarity (e.g. word2vec~\cite{mikolov2013distributed} cosine similarity). Relational Similarity (RS) selects indicator entities as those most likely to satisfy triples connecting current entities to the new entity through relations. Hybrid Similarity (ERS) combines both algorithms by first selecting an initial indicator set using ES, then filtering to the final indicator entities using RS. These methods directly generalize to other multi-relational embedding types (e.g. TransE~\cite{bordes2013transe}, Complex~\cite{trouillon2016complex}) because they rely on identifying an indicator set of entities without making assumptions about the multi-relational embedding type.

\textbf{Entity Similarity (ES) Initialization} leverages word2vec \cite{mikolov2013distributed} to select the indicator entities because word2vec captures distributed semantics of words, which helps identify contextually similar entities. The indicator set $\mathcal{I}$ comprises of known entities $e\!\in\!\mathcal{E}^{n-1}\,$that have the highest cosine similarity between their word2vec vector $\pmb{\pi}_{e}$ and the OOKB entity's word2vec vector $\pmb{\pi}_{\hat{e}}$, in Equation~\ref{eq:similarity}.
%The set of entities that have the highest cosine similarity with the OOKB entity are selected as the indicator entities by comparing each entity's corresponding word2vec vector $v^j_{\mathcal{E}w}$ with the OOKB entity's word2vec vector $v^{|\mathcal{E}^{n-1}|+1}_{\mathcal{E}w}$ as in Equation~\ref{eq:similarity} below.

\vspace{-0.5cm}
\begin{equation}
	\begin{aligned}
		\mathcal{I}_{\textrm{ES}} = \argtopk_{e\in\mathcal{E}^{n-1}}\big(\pmb{\pi}_{e} \cdot \pmb{\pi}_{\hat{e}}\big)
 	\end{aligned}
 	\label{eq:similarity}
\end{equation}
\vspace{-0.5cm}

Where $\argtopk$ selects top $k=|\mathcal{I}_{\textrm{ES}}|$ entities with the highest scores. The OOKB entity vector $\textbf{v}_{\hat{e}}^n$ is then initialized as the centroid of the vectors in $\mathcal{I}_{\textrm{ES}}$ (Equation~\ref{eq:indicator}). Figure~\ref{fig:approach} shows a diagram for ES initialization where the triangles are indicator entities.

\textbf{Relational Similarity (RS) Initialization} selects the indicator entities $\mathcal{I}$ using a set of insert triples $\{(h,r,t)_i\}$ that connect the new entity to current entities in the embedding via relations (i.e. the triples must satisfy $h_{i}\in\mathcal{E}^{n-1}$ if $t_{i}=\hat{e}$ or $t_{i}\in\mathcal{E}^{n-1}$ if $h_{i}=\hat{e}$ so that it can inform initialization). The insert triples would be observed by the robot when an OOKB entity is encountered.

%After grouping triples by relation type, a resultant vector $\pmb{\alpha}_{r,j}$ for each triple is computed from the known parameters of the triple (i.e. possible location for $\textbf{v}_{\hat{e}}$ based on a single triple). Therefore, when $h_{j} = \hat{e}$, $r_{j}$ and $t_{j}$ are used to compute the resultant vector, while when $t_{j} = \hat{e}$, $r_{j}$ and $h_{j}$ are used. Equation~\ref{eq:resultants} shows this procedure for ANALOGY.
For each relation type $r$, resultant vectors are computed from the subset of insert triples with that relation type $\{(h_j,r,t_j)\}\subset\{(h,r,t)_i\}$. These resultant vectors infer the possible locations of the OOKB entity based on a single triple. Specifically, for each triple in the subset, $\pmb{\alpha}_{r,j}$ is computed from known parameters of $r$ and $h_{j}$ when $t_{j} = \hat{e}$ or $r$ and $t_{j}$ when $h_{j} = \hat{e}$. Equation~\ref{eq:resultants} shows this procedure for ANALOGY.

\vspace{-0.5cm}
\begin{equation}
	\begin{aligned}
		\pmb{\alpha}_{r,j} = 
		\bigg\{&
		\begin{tabular}{l}
		$(\textbf{v}^{n-1}_{h_j})^T \cdot {\textbf{W}_{r}}$ \quad\;\,
		if $h_{j}\in\mathcal{E}^{n-1}$ and $t_{j}=\hat{e}$\\[4pt]
		$(\textbf{v}^{n-1}_{t_j})^T \cdot {\textbf{W}^{-1}_{r}}$ \quad\:\!\!
		if $h_{j}=\hat{e}$ and $t_{j}\in\mathcal{E}^{n-1}$ \\
		\end{tabular}
	\end{aligned}
	\label{eq:resultants}
\end{equation}
\vspace{-0.3cm}

All the resultant vectors for each relation type $r$ are combined by averaging to get resultant vector centroids $\bar{\pmb{\alpha}}_r$. The set of entities that have the highest accumulated cosine similarities to each resultant vector centroid (i.e. across each relation type) are selected as the indicator entities, as shown below.

\vspace{-0.7cm}
\begin{equation}
	\begin{aligned}
		\mathcal{I}_{\textrm{RS}}=
		\argtopk_{e\in\mathcal{E}^{n-1}}\Bigg(\sum_{r\in\mathcal{R}^{n-1}} \textbf{v}_{e}\cdot\bar{\pmb{\alpha}}_r\Bigg)
	\end{aligned}
	\label{eq:relational}
\end{equation}
\vspace{-0.3cm}

The initial value of $\textbf{v}_{\hat{e}}^n$ is then the centroid of the selected indicator set vectors $\mathcal{I}_\textbf{RS}$ as in Equation~\ref{eq:indicator}. Figure~\ref{fig:approach} shows the conceptual diagram for RS initialization where the triangles are indicator entities.

\textbf{Hybrid Similarity (ERS) Initialization} is informed by entity similarities as well as relations between entities by combining the two previous algorithms. First, a preliminary indicator set $\mathcal{I}_{\textrm{ES}}$ of the most similar entities is selected using the ES algorithm, as in Equation~\ref{eq:similarity}. The entities in $\mathcal{I}_{\textrm{ES}}$ are then used as inputs to the RS algorithm by requiring entities in $\argtopk$ of Equation~\ref{eq:relational} to be in $\mathcal{I}_{\textrm{ES}}$. The RS algorithm further filters the preliminary indicator entities $\mathcal{I}_{\textrm{ES}}$ to select the subset of entities that most likely satisfy the set of insert triples. The set of entities output by RS are the final set of entities that become the indicator set:

\vspace{-0.4cm}
\begin{equation}
	\begin{aligned}
		\mathcal{I}_{\textrm{ERS}}=
		\argtopk_{e\in\,\mathcal{I}_{\textrm{ES}}}\Bigg(\sum_{r\in\mathcal{R}^{n-1}} \textbf{v}_{e}\cdot\bar{\pmb{\alpha}}_r\Bigg)
	\end{aligned}
	\label{eq:hybrid}
\end{equation}
\vspace{-0.3cm}

%===============================================================================
\vspace{-0.4cm}
\section{Experimental Settings}
\label{sec:settings}
\vspace{-0.2cm}

Our evaluation is inspired by a learning scenario in which a service robot incrementally acquires novel semantic knowledge about objects in its environment. We obtain our knowledge graph by mining AI2Thor \cite{kolve2017ai2}, a highly realistic simulator of household environments, which enables us to capture the diverse nature of real-world environments. 
%Note that our data includes many-to-many relationships between entities (a single object can be at many possible locations, and can possess many possible properties), unlike many other works on knowledge graph embedding \cite{multiple,examples,here} that utilize datasets with one-to-one mappings \SC{?or is it one-to-many?} (WordNet \cite{}, Freebase \cite{}).  
Below, we describe our data, performance metrics, experimental procedure, and parameters.

%As a result, our experimental settings differ from standard knowledge graph embedding works (see Section~\ref{sec:related}) because our models need to capture the statistical nature of real-world environments, described in Section~\ref{sec:settings_graph}. In addition, our modified performance metrics, also in Section~\ref{sec:settings_graph}, are more compatible with the relaxed assumptions not accounted for in the standard metrics. We conclude with Section~\ref{sec:settings_procedure} that describes our experimental procedure to simulate an incremental learning scenario, and Section~\ref{sec:settings_details} detailing all parameter settings required to recreate experimental results.\zk{The latter part of the paragraph, which is really an outline, could probably be cut for space}

\vspace{-0.3cm}
\subsection{Knowledge Graph \& Metrics}
\label{sec:settings_graph}
\vspace{-0.2cm}

The knowledge graph used in this work was mined from AI2Thor, a realistic home simulator (see Table~\ref{tbl:thor}).  We utilize this data because benchmark datasets commonly used widely across multi-relational embedding works~\cite{wang2017kge_survey,bordes2013transe,liu2017analogical} do not capture the statistical nature of real-world encountered by service robots. In particular, both Freebase~\cite{bollacker2008freebase} and WordNet~\cite{miller1995wordnet} only contain unique triples of factual information (e.g. {\small$($\textit{cup, hypernym, container}$)$}, {\small$($\textit{StevenSpeilberg, directorOf, Jaws}$)$}); however, using distributions of non-unqiue triples more closely models the real-world due to variance between environments.

\begin{wraptable}{r}{0.46\textwidth}
    \vspace{-.8cm}
	\caption{AI2Thor Knowledge Graph Statistics}
	\centering      
	\begin{tabular}{@{}lccccc@{}} \toprule
		\multicolumn{6}{c}{\small 3 Relation Types, 106 Entities} \\ \midrule
		\multicolumn{6}{c}{\small Median Count per Environment} \\ \midrule
		\makecell[c]{\small Room \\ \small Type} & \makecell[c]{\small Loc. \\ \small Rel.} & \makecell[c]{\small Mat. \\ \small Rel.} & \makecell[c]{\small Aff. \\ \small Rel.} & \makecell{\small Num. \\ \small Ent.} & \makecell[c]{\small Num. \\ \small Rooms} \\ \midrule
		\small Bath- & 28 & 21 & 46 & 18 & 30 \\ 
		\small Bed- & 28.5 & 16 & 54.5 & 20 & 30 \\ 
		\small Kitchen & 59.5 & 51 & 109 & 27 & 30 \\ 
		\small Living- & 22.5 & 8 & 37 & 20 & 30 \\ 
		\small All & 29.5 & 18.5 & 50 & 20 & 120 \\ \bottomrule
	\end{tabular}
	\label{tbl:thor}
	\vspace{-0.5cm}
\end{wraptable}

We manually extended the set of AI2Thor entities, comprising 82 household concepts (e.g. microwave, toilet, kitchen) and 17 affordances (e.g. pick up, open, turn on), to include 7 material properties (e.g. wood, fabric, glass), which were assigned probabilistically based on materials encountered in the SUNCG dataset \cite{song2016ssc} for a total of 106 entities. In total, our dataset contains over 15K triples, of which 352 are unique. Many triples are repeated according to distributions of the default AI2Thor environments (e.g., {\small\textit{(bowl, atLocation, cabinet)}} occurs 22 times).
 
Responses to queries about the AI2Thor knowledge graph are best quantified on a scale because of the uncertain nature of realistic environments (e.g. multiple potential locations are likely for a given object with varying likelihoods). As a result, ground truth responses are ranked lists of candidates ordered according to observations of a unique triple in all default environments of AI2Thor (i.e. more observations give higher ranks). Instead of mean-reciprocal-rank (MRR) over a set of $N$ queries in Equation~\ref{eq:mrr} that assumes a ground truth rank of 1 \cite{wang2017kge_survey,bordes2013transe,liu2017analogical}, we report MRR* in Equation~\ref{eq:mrr*} that supports variable ground truth ranks by including a ground truth rank variable $R_G^n$ in addition to the predicted rank $R_P^n$.

\vspace{-0.2cm}
\noindent\begin{minipage}{.5\linewidth}
    \begin{equation}
    \textrm{MRR} = \frac{1}{N}\sum_{n=1}^{N} \frac{1}{R_P^n}
    \label{eq:mrr}
    \end{equation}
\end{minipage}
\begin{minipage}{.5\linewidth}
    \begin{equation}
    \textrm{MRR*} = \frac{1}{N}\sum_{n=1}^{N} \frac{1}{\mid R_G^n-R_P^n \mid+1}
    \label{eq:mrr*}
    \end{equation}
\end{minipage}

\vspace{-0.3cm}
\subsection{Experimental Procedure}
\label{sec:settings_procedure}
\vspace{-0.2cm}

To model an incremental learning scenario across all experiments, we first learn an initial embedding $\Theta^{0}$ from an initial dataset $\mathcal{D}^{0}$. Then $\Theta^{1}$, which is trained on a second dataset $\mathcal{D}^{1}$, is initialized by reusing embeddings from $\Theta^{0}$ and inserting OOKB entities $\xi^0$ using an initialization method. Only two learning sessions were used in each experimental case because each initialization method can train to convergence with enough epochs, making a third learning session equivalent to restarting at the first learning session.

$\mathcal{D}^{0}$ consists of train, validation, and test sets of distinct unique triples (i.e. $\mathcal{D}^{0}_{Tr} \cap (\mathcal{D}^{0}_{Va} \cup \mathcal{D}^{0}_{Te}) = \mathcal{D}^{0}_{Va} \cap \mathcal{D}^{0}_{Te} = \emptyset$). $\mathcal{D}^{0}$ is limited to only triples related to known entities $\mathcal{E}^{0}$ while all triples related to OOKB entities $\xi^{0}$ are withheld. The second dataset $\mathcal{D}^{1}$ contains triples related to all entities including $\xi^{0}$, so that $\mathcal{E}^{1} = \mathcal{E}^{0} \cup \xi^{0}$. Therefore, datasets generated for the later session of incremental learning subsume previous datasets. Before beginning the second training session, embeddings $\Theta^1 = \big\{\{\textbf{v}^1_{e}|\,e\in\mathcal{E}^1\},\{\textbf{W}^1_{r}|\,r\in\mathcal{R}^1\}\big\}$ are initialized using to Equations~\ref{eq:ent_init} and~\ref{eq:rel_init} below, where \textbf{ookb\_init} is one of the proposed (Section~\ref{sec:approach}) or baseline (Section~\ref{sec:settings_details}) initialization algorithms.

\vspace{-0.2cm}
\noindent\begin{minipage}{.5\linewidth}
\begin{equation}
	\textbf{v}^{1}_{e}\!= 
	\! \bigg\{ \!
	\begin{tabular}{l}
	$\textbf{v}^{0}_{e} \qquad\qquad\quad\:\:\forall\,e\in{\mathcal{E}}^{0}$ \\
	$\textbf{ookb\_init}(e) \quad\:\forall\,e\in{\xi}^{0}$
	\end{tabular}
\label{eq:ent_init}
\end{equation}
\end{minipage}
\begin{minipage}{.5\linewidth}
\begin{equation}
	\begin{aligned}
	\textbf{W}^1_{r} \! = \textbf{W}^0_{r} \, \forall \, r\in\mathcal{R}^{1}
	\end{aligned}
\label{eq:rel_init}	
\end{equation}
\end{minipage}
\vspace{-0.2cm}

\textit{Fine-tuning} was used to train the second model's parameters $\Theta^{1}$ over $\mathcal{D}^{1}$ because it has a simple implementation and our contribution is not focused on the catastrophic-forgetting problem. Additionally, results in~\cite{song2018enriching} using better approaches like EWC~\cite{kirkpatrick2017overcoming} were only marginally better than fine-turning (2.08\%). In fine-tuning, learning rates are lowered during incremental learning sessions but no new training regularization is included.

\vspace{-0.2cm}
\subsection{Parameter Details \& Baselines}
\label{sec:settings_details}
\vspace{-0.2cm}

Throughout our experiments we measure and log the MRR* at each epoch when learning over dataset $\mathcal{D}^{1}$ until convergence, explicitly controlling all other variables to allow direct comparisons between different initialization methods. Convergence\footnote{The convergence condition is when the MRR* is within 8 MRR* of the joint-learning model performance.} was determined using a \textit{joint-learning} (Joint) model as in \cite{song2018enriching}, which is essentially a batch learned multi-relational embedding trained only on $\mathcal{D}^{1}$ serving as an upper-bound.

The two baselines used in our experiments initialize new entity embeddings uniformly distributed over ranges determined by different criteria. Normalized-initialization (Xavier), used in \cite{song2018enriching}, is uniformly distributed based on the dimensionality of the vector $d_\mathcal{E}$ so that the minimum value for each $j$ dimension is $v^j_{min} = -\sfrac{6}{\sqrt{d_\mathcal{E}}}$ and the max value is $v^j_{max} = \sfrac{6}{\sqrt{d_\mathcal{E}}}$. The other baseline, we termed informed-uniform (IU), is uniformly distributed based on the range of all current entity embeddings $\textbf{v}^0_{e}\,\forall\,e\in\mathcal{E}^0$ so that the minimum value for each $j$ dimension $v^{j}_{min}=min(\textbf{v}^0_{e})$ and the max value is $v^{j}_{max}=max(\textbf{v}^0_{e})$. Equation~\ref{eq:uniform_init} shows how both baselines initialize entity embeddings.

\vspace{-0.3cm}
\begin{equation}
	\begin{aligned}
		\textbf{v}^1_{e,j}\! = \!
		\bigg\{\!
		\begin{tabular}{l}
			$\textbf{v}^{0}_{e,j}$ \qquad\qquad\quad\,\;
			if $e\in\mathcal{E}^{0}$ \\
			$U(v^{j}_{min},v^{j}_{max})$ \quad 
			if $e\in\mathcal{\xi}^{0}$ \\
		\end{tabular}
	\end{aligned}
	\label{eq:uniform_init}
\end{equation}
\vspace{-0.3cm}

We determined that the best dimensionality for vectors and mappings was 100, ratio of negative over positive samples was 9, and learning rate and weight decay to train $\Theta^{0}$ was $1\mathrm{e}{-1}$ and $1\mathrm{e}{-3}$, respectively, for all experiments using cross-validation when training the joint-learning model. When training $\Theta^{1}$ (i.e. fine-tuning), the maximum number of epochs allowed was 150, and the learning rate was decreased to $2\mathrm{e}{-3}$. All results are reported in a `filtered' setting~\cite{bordes2013transe}, where triples already within that training and validation sets are removed before ranking. The set of OOKB entities $\xi^{0}$ in each experimental case are uniformly randomly selected as in~\cite{song2018enriching}. This is repeated 30 times for each size of OOKB entity set $|\xi^{0}|\!\in\!{1,...,10}$, recording the sets of $\xi^{0}$ so they match across initialization methods.

\begin{figure}[t]
	\centering
	\begin{subfigure}[b]{0.49\columnwidth}
		\centering 
    	\includegraphics[width=1.0\textwidth]{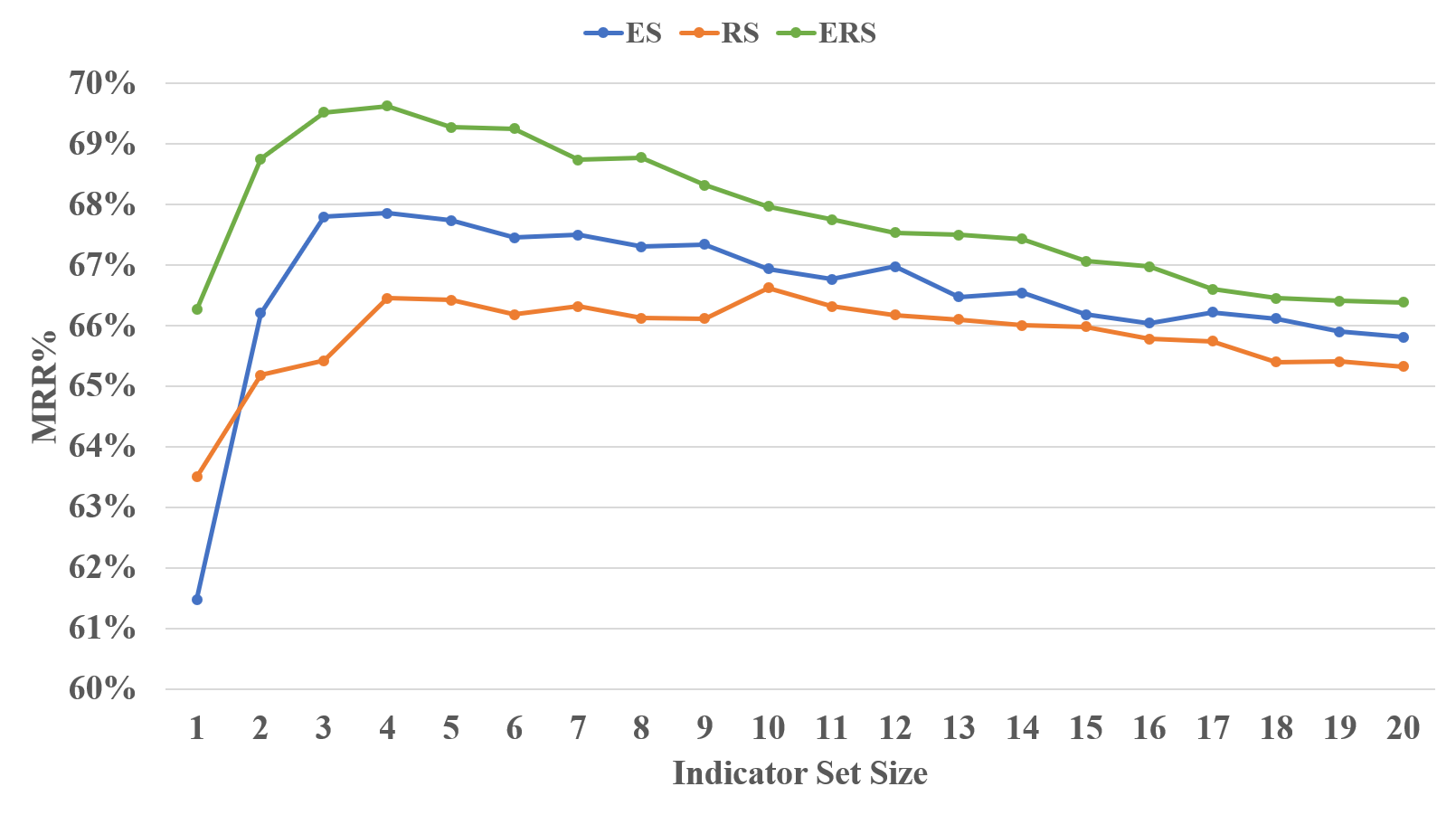}
    	\caption[]%
    	{{}}
    	\label{fig:hp_query}
		\vspace{-0.2cm}
	\end{subfigure}%
	~
	\begin{subfigure}[b]{0.49\columnwidth}
		\centering 
    	\includegraphics[width=1.0\textwidth]{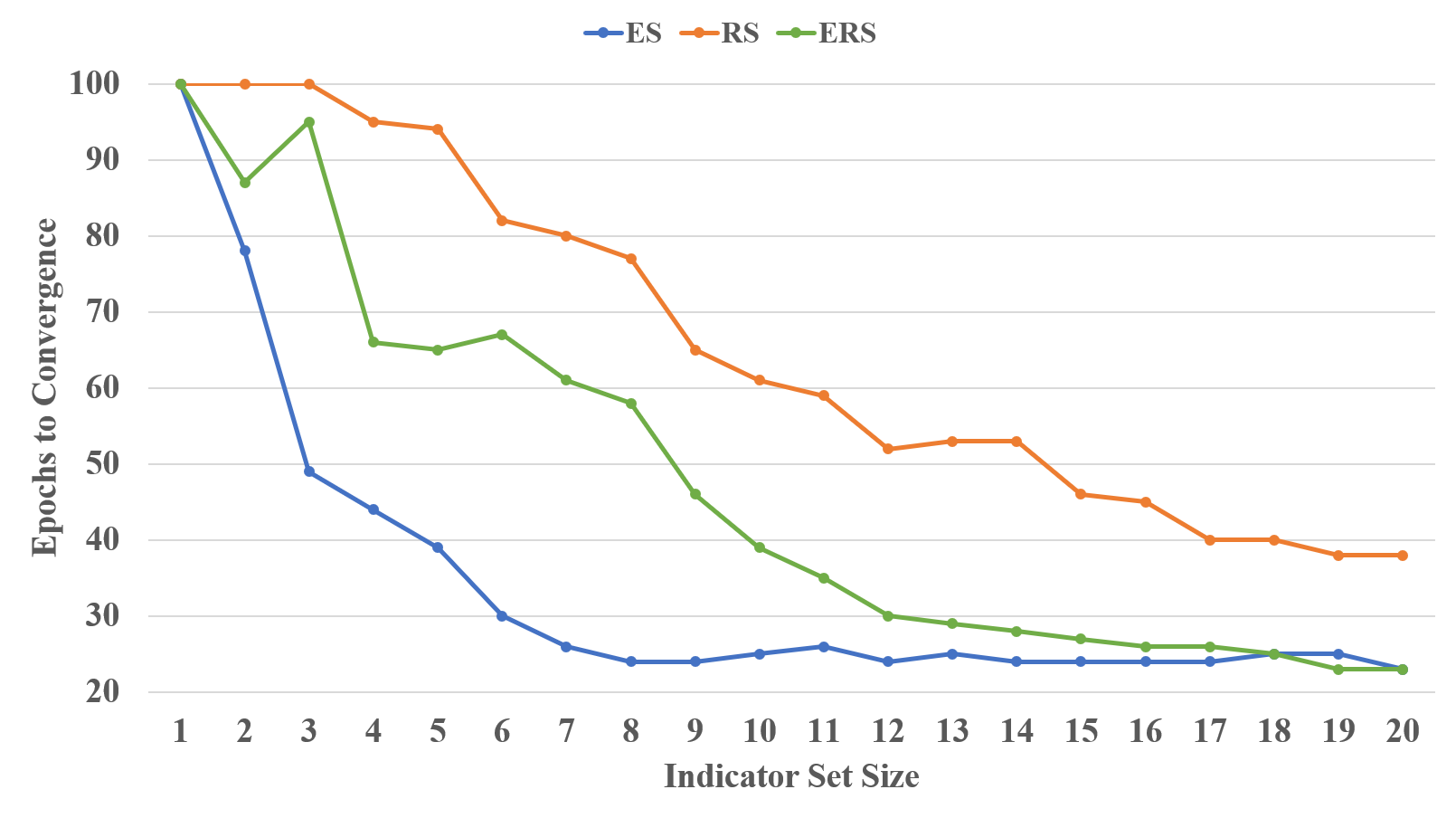}
    	\caption[]%
    	{{}}
    	\label{fig:hp_converge}
		\vspace{-0.2cm}
	\end{subfigure}
	\caption[]%
	{\small(a)~Immediate MRR*\% sensitivity to indicator set size hyper-parameter, and (b)~Epochs-to-Convergence sensitivity to indicator set size hyper-parameter.}
	\vspace{-0.7cm}
\end{figure}

To determine the best indicator set size for each initialization algorithm, we ran a hyper-parameter sensitivity analysis considering MRR* and convergence seen in Figures~\ref{fig:hp_query} and \ref{fig:hp_converge}, respectively. After size 4, as indicator set size is increased, MRR* performance degrades while convergence improves. Noticing this trade-off, for each algorithm we increased the indicator set size while the average number of epochs for three neighboring sizes decreased by 1 epoch or the MRR* went 1\% point below the best performance, leading to indicator entity set sizes of 8, 18, and 9 for ES, RS, and ERS algorithms, respectively. Additionally, ERS used an initial indicator entity set size of 30.

% FIGURE HP
% FIGURE HP

\vspace{-0.4cm}
%===============================================================================
\section{Experimental Results}
\label{sec:results}
\vspace{-0.2cm}

To better understand the different initialization methods, our experiments probe how each affects the immediate inference performance (Section~\ref{sec:results_query}), the time-to-convergence measured in epochs (Section~\ref{sec:results_convergence}), and the quality of knowledge association (Section~\ref{sec:results_mp3d}). Here, the quality of knowledge association refers to how well new entities initialized with each method integrate with inferences about previous entities.
%Across all experiments, the procedure was to learn an initial embedding, insert new entities into the embedding, measure immediate performance, then begin additional training to optimize for the new entities, logging measurements of performance.

\vspace{-0.2cm}
\subsection{Improved Immediate Inferences}
\label{sec:results_query}
\vspace{-0.2cm}

Concepts added to robot knowledge representations should be initialized to semantically meaningful values, enabling more accurate immediate inferences because deployed robots often must reason about new concepts without enough time to optimize their learning models for the newly encountered concepts. To evaluate each initialization method regarding this criterion, we learned an initial multi-relational embedding over a subset of the entities in the AI2Thor dataset, then initialized new OOKB entities in the embedding using each initialization method, and measured the inference performance before additional training.

% FIGURE THOR
\begin{figure}[t]
	\centering
	\begin{subfigure}[b]{0.49\columnwidth}
		\centering
    	\includegraphics[width=1.0\textwidth]{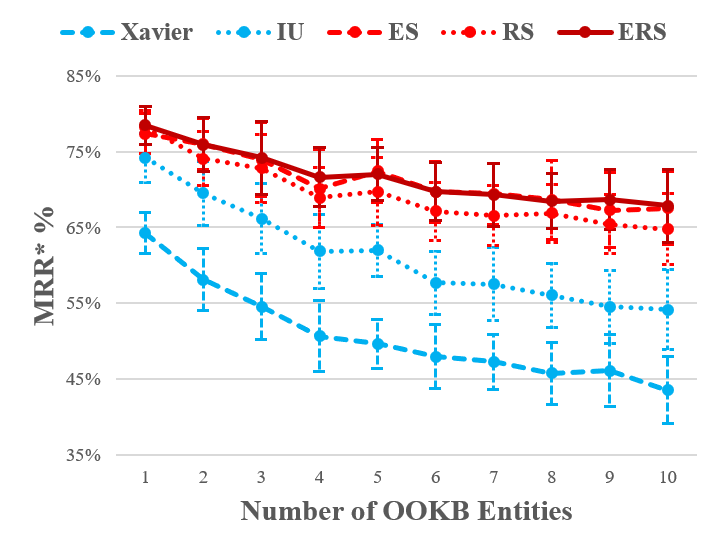}
		\caption[]%
		{{}}
		\label{fig:thor_query}
		\vspace{-0.2cm}
	\end{subfigure}%
	~
	\begin{subfigure}[b]{0.49\columnwidth}
		\centering
        \includegraphics[width=1.0\textwidth]{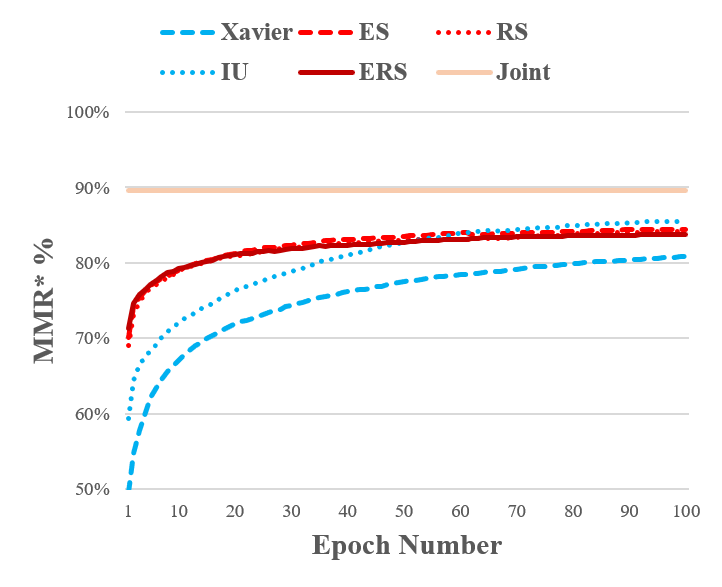}
        \caption[]%
		{{}}
        \label{fig:thor_convergence}
		\vspace{-0.2cm}
	\end{subfigure}
	\caption[]
	{\small (a)~ISI outperforms Xavier initialization across all AI2Thor queries before additional training, and (b)~Xavier initialization requires more epochs to converge than ISI during incremental learning sessions.}
	\vspace{-0.5cm}
\end{figure}
% FIGURE THOR

Figure~\ref{fig:thor_query} reports the average MRR* performance using each entity initialization method before performing additional training to optimize the second model (i.e. $\Theta^{1}$). Each point is the weighted average MRR* across all queries in $\mathcal{D}^{1}$ measured for an entity initialization method and OOKB entity set size ranging from roughly 1\% to 10\% of $|\mathcal{E}|$, while keeping initial embeddings and other variables the same across initialization methods.

In Figure~\ref{fig:thor_query}, we see that across various sizes of new entities being inserted (i.e. $|\xi^{0}|$), ISI methods give better inference results for queries. Across all sizes of $\xi^{0}$, ES, RS, and ERS initialization out perform  Xavier initialization by an average 41.4\%, 37.6\%, 42.1\%, respectively. Therefore, on average, vectors for new concepts initialized with ISI give better inference results than baselines and hence are more semantically meaningful initial embedding vectors.

\vspace{-0.4cm}
\subsection{Decreased Epochs-to-Convergence}
\label{sec:results_convergence}
\vspace{-0.2cm}

\begin{wraptable}{r}{0.23\textwidth}
    \vspace{-1cm}
    \centering
    \captionof{table}{\small Epochs-to\\-Convergence}
    \begin{tabular}{@{}lcc@{}} \toprule
        \makecell{\small Init. \\\small Method} &\small Avg. & \makecell{\small Std. \\\small Dev.} \\ \midrule
        \small Xavier & 112.6 &\small 48.1 \\
        \small IU &\small 37.9 &\small 16.2 \\
        \small ES &\small 16.6 &\small 9.9 \\
        \small RS &\small 27.1 &\small 17.5 \\
        \small ERS &\small 29.9 &\small 23.3 \\ \bottomrule
    \end{tabular}
    \label{tbl:thor_convergence}
    \vspace{-0.7cm}
\end{wraptable}

Concepts added to robot knowledge representations should be efficiently integrated to save computations on deployed robots that are often compute resource and time deprived. To evaluate each initialization method regarding this criterion, we learned an initial multi-relational embedding over a subset of the entities in the AI2Thor dataset, then initialized new OOKB entities in the embedding using each initialization method, and measured how many epochs were required to converge within 8\% of the joint-learning model performance during additional training.

Figure~\ref{fig:thor_convergence} shows the average MRR* performance across all queries in $\mathcal{D}^{1}$ during the second learning session where $|\xi^{0}|=5$. At each epoch, the current weighted average MRR* for all queries in $\mathcal{D}^{1}$ is logged to generate the learning curve for each initialization method. This is repeated for each $\xi^{0}$ size from the experiment in Section~\ref{sec:results_query}, resulting in the averages and standard deviations of Table~\ref{tbl:thor_convergence}.

Table~\ref{tbl:thor_convergence} shows that across various sizes of new OOKB entities being inserted (i.e. $|\xi^{0}|$), ISI methods converge faster than Xavier on average. Across all sizes of $\xi^{0}$, ES, RS, and ERS initialization on average require 85.3\%, 75.9\%, and 73.4\%, respectively, fewer epochs to converge than Xavier initialization. Therefore, ISI helps to reduce the number of computations required to optimize a multi-relational embedding with newly initialized concepts.

\vspace{-0.4cm}
\subsection{Mitigated Knowledge Corruption}
\label{sec:results_mp3d}
\vspace{-0.2cm}

In addition to accuracy of semantic meaning and efficiency of integration, concepts added to robot knowledge representations should also associate well with current knowledge, mitigating corruption to previously learned concepts. To test this property with each initialization method, we considered a common situation where a robot first learns an embedding in simulation (i.e. AI2Thor), then gets deployed to a realistic environment encountering new concepts (i.e. MatterPort3D (MP3D)~\cite{chang2017matterport3d}), and requires the set of known entities to be extended.

To model this sim-to-real scenario, an initial multi-relational embedding of all entities in AI2Thor was learned, and subsequently 10 new entities from a subset of MP3D were initialized in the embedding using each initialization method. Finally, the MRR* performance with respect to only entities in AI2Thor was logged during additional training. The procedure was repeated 30 times to generate all results with the subset of MP3D (i.e. 50 entities, randomly selected and filtered for a minimum of 6 non-unique triples).

Following testing procedures from Sections~\ref{sec:results_query} and \ref{sec:results_convergence}, we first probed each initialization method for accuracy and efficiency when initializing entities from MP3D and found ISI to outperform Xavier initialization. When learning new concepts across datasets, ISI improves immediate inference performance (Xavier, IU, ES, RS, ERS performed with 50.6, 77.3, 82.1, 80.9, 82.5 MRR*, respectively) and speeds up time-to-convergence (only Xavier required on average 90 epochs to converge when inserting new concepts).

\begin{wrapfigure}{r}{0.5\textwidth}
    \vspace{-0.6cm}
	\centering 
	\includegraphics[width=0.5\textwidth]{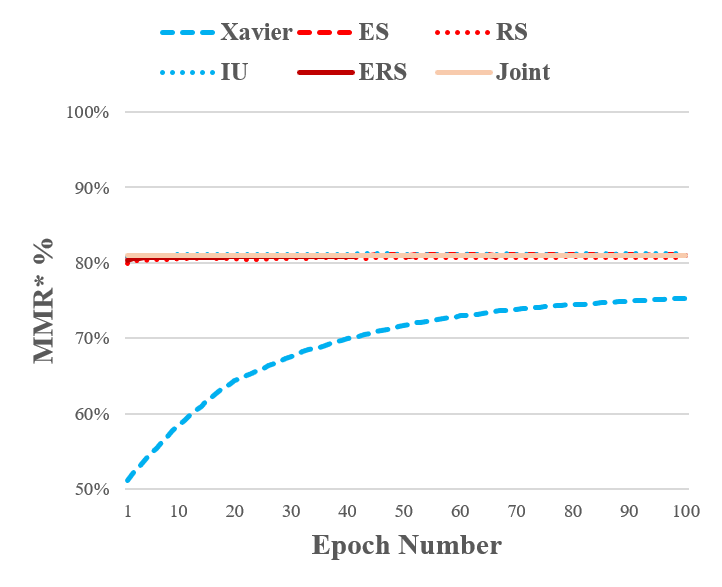}
	\caption[]
	{\small ISI causes negligible corruption to previous embeddings when compared to Xavier initialization.}
	\label{fig:mp3d_convergence_g}
	\vspace{-0.5cm}
\end{wrapfigure}

In addition, the experiments showed that ISI mitigated corruption to AI2Thor embeddings when initializing MP3D entities. In Figure~\ref{fig:mp3d_convergence_g} the MRR* performance with respect to only entities in AI2Thor was logged for each initialization method during the additional training that included MP3D data. Xavier was the only initialization method to have significant effects on inference performance, dropping immediate MRR* over triples related only to AI2Thor by 37.0\%. Similar results were experienced within the AI2Thor experiments, but the observation was only highlighted here because the distinct datasets make the explanation clear.

% FIGURE MP3D CORRUPTION
% FIGURE MP3D CORRUPTION

Initializing new entity embeddings with Xavier likely reduces MRR* over previous concepts because new entities from MP3D are forced into parts of the embedding space disparate from their semantic meaning, drastically changing during additional training. ISI mitigates this misplacement by adding MP3D entities to semantically similar regions of the AI2Thor entity embedding space\footnote{Note that while IU also did not cause drastic corruptions to AI2Thor embeddings, it performed more poorly when making queries regarding only newly inserted concepts from MP3D (Xavier, IU, ES, RS, ERS performed with 50.2, 67.8, 85.7, 83.6, 88.4 MRR*, respectively).}.

Mitigating prior knowledge corruption by using ISI also enables multi-relational embeddings to make better generalizations about new concepts using semantic similarities to previous concepts. Therefore, MP3D entities inserted into a multi-relational embedding originally learned from AI2Thor can immediately receive more reasonable rankings of affordances, despite having only ``\textit{atLocation}'' relations for those entities. Affordance rankings for several of the MP3D entities added to the AI2Thor multi-relational embedding are shown in Table~\ref{tbl:aff_ranks} where generalizations in red are semantically incorrect and others in yellow are highly unlikely.

\begin{table}[t]
    \vspace{-0.7cm}
	\caption{Ranked Affordance Generalizations for Entities in MatterPort3D}
	\centering      
	\begin{tabular}{@{}l|l|l|l|l|l@{}} \toprule
		\multicolumn{2}{c}{\textit{fan} entity} & \multicolumn{2}{|c|}{\textit{bottle} entity} & \multicolumn{2}{c}{\textit{stove} entity}\\ \midrule
		\multicolumn{1}{c|}{\small{ISI}} & \multicolumn{1}{c|}{\small{Xavier}} & \multicolumn{1}{c|}{\small{ISI}} & \multicolumn{1}{c|}{\small{Xavier}} & \multicolumn{1}{c|}{\small{ISI}} & \multicolumn{1}{c}{\small{Xavier}} \\ \midrule
		\small{1. pick up (v)} & \small{1. pick up (v)} & \small{1. put (v)} & \small{\color{red}1. vase (n)} & \small{1. open (v)} & \small{\color{red}1. shelf (n)} \\
		\small{2. put (v)} & \small{\color{red}2. stone (n)} & \small{2. pick up (v)} & \small{\color{red}2. shelf (n)} & \small{\color{orange}2. pick up (v)} & \small{\color{red}2. vase (n)} \\
		\small{3. turn on (v)} & \small{\color{red}3. glass (n)} & \small{3. fill (v)} & \small{3. pick up (v)} & \small{3. turn off (v)} & \small{3. turn off (v)} \\
		\small{4. turn off (v)} & \small{\color{orange}4. empty (v)} & \small{\color{orange}4. slice (v)} & \small{4. break (v)} & \small{4. close  (v)} & \small{\color{red}4. painting (n)} \\ \bottomrule
	\end{tabular}
	\label{tbl:aff_ranks}
	\vspace{-0.5cm}
\end{table}

%===============================================================================
\vspace{-0.3cm}
\section{Conclusion}
\label{sec:conclusion}
\vspace{-0.2cm}

We presented Incremental Semantic Initialization as a means of adding OOKB entities to a previously learned embedding, as a result enabling the practical use of multi-relational embeddings in incremental robot learning scenarios.  The ISI techniques, which reason about the current embedding space to initialize new embeddings, more efficiently and accurately initialize new concepts than previous methods used in~\cite{song2018enriching}, while mitigating corruption of previous concepts. The accompanying video demonstrates the application of this work to a physical robot learning scenario.

%===============================================================================

% The maximum paper length is 8 pages excluding references and acknowledgements, and 10 pages including references and acknowledgements

\clearpage
% \acknowledgments{If a paper is accepted, the final camera-ready version will (and probably should) include acknowledgments. All acknowledgments go at the end of the paper, including thanks to reviewers who gave useful comments, to colleagues who contributed to the ideas, and to funding agencies and corporate sponsors that provided financial support.}

%===============================================================================

\bibliography{references}  % .bib

\end{document}